# Lost and Found:
# Detecting Small Road Hazards for Self-Driving Vehicles

Peter Pinggera[1,3,*], Sebastian Ramos[1,2,*], Stefan Gehrig[1], Uwe Franke[1], Carsten Rother[2], Rudolf Mester[3]

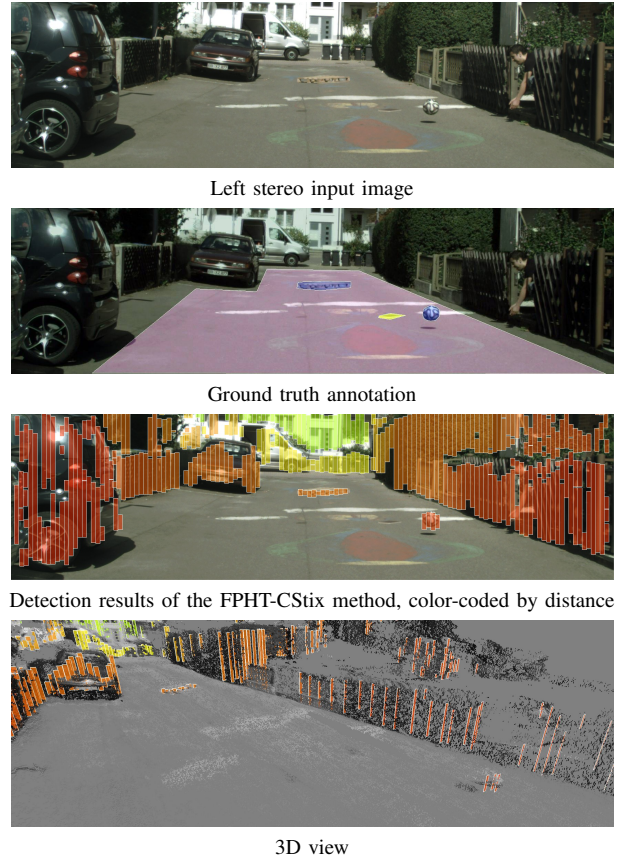

Fig. 1. Exemplary scene from the *Lost and Found* dataset and corresponding results of the proposed approach. The setup includes static and moving hard-to-detect obstacles such as a EUR-pallet and a soccer ball (blue annotations), as well as a non-hazardous flat piece of wood (yellow annotation). Ground truth free-space is shown in purple

*Abstract*—Detecting small obstacles on the road ahead is a critical part of the driving task which has to be mastered by fully autonomous cars. In this paper, we present a method based on stereo vision to reliably detect such obstacles from a moving vehicle.

The proposed algorithm performs statistical hypothesis tests in disparity space directly on stereo image data, assessing free-space and obstacle hypotheses on independent local patches. This detection approach does not depend on a global road model and handles both static and moving obstacles.

For evaluation, we employ a novel lost-cargo image sequence dataset comprising more than two thousand frames with pixel-wise annotations of *obstacle* and *free-space* and provide a thorough comparison to several stereo-based baseline methods. The dataset will be made available to the community to foster further research on this important topic[4].

The proposed approach outperforms all considered baselines in our evaluations on both pixel and object level and runs at frame rates of up to 20 Hz on 2 mega-pixel stereo imagery. Small obstacles down to the height of 5 cm can successfully be detected at 20 m distance at low false positive rates.

## I. INTRODUCTION

The detection of small obstacles or debris on the road is a crucial task for autonomous driving. In the US, 25000 crashes per year due to road debris were reported in 2004 [1], and approximately 150 people were killed by accidents involving lost hazardous cargo in 2011 [2]. In Austria, a dedicated initiative was launched in 2006 to detect hazardous cargo in tunnels to improve traffic safety [3].

The task of detecting small but potentially hazardous cargo on the road proves to be quite difficult, even for experienced human drivers (see e.g. Fig. 1). Different sensor types can be applied to the problem, from passive cameras to active radar or lidar sensors. While active range sensors provide high accuracy in terms of point-wise distance and velocity measurement, they typically suffer from low resolution and high cost. In contrast, cameras provide very high spatial resolution at relatively low cost. However, the detection task of small obstacles is a very challenging problem from a computer vision perspective, since the considered objects cover very small image areas and come in all possible shapes and appearances.

Stereo or multi-camera setups allow for the computation of dense range maps of the observed environment, being increasingly popular for application in self-driving cars and mobile robots in general. Unfortunately, an inherent drawback of stereo vision systems is the comparatively low accuracy of distance measurements, especially at long ranges. However, accuracy is crucial for the timely detection of small obstacles and an appropriate response by safety-critical moving platforms.

In this paper, we build upon previous work of [4] and extend it in two ways: We present a reparametrization of the underlying problem to boost efficiency, achieving a speed up of factor 10 while keeping the quality of the results at the highest level. In addition, inspired by the established Stixel representation [5], we introduce a mid-level obstacle repre-sentation based on the original point-based output, resulting in improved robustness and compactness that significantly aids further processing steps.

[1]Environment Perception, Daimler R&D, Sindelfingen, Germany
{firstname.lastname}@daimler.com
[2]Computer Vision Lab Dresden, TU Dresden, Germany
carsten.rother@tu-dresden.de
[3]VSI Lab, Computer Science Dept., Goethe Univ., Frankfurt, Germany
mester@vsi.cs.uni-frankfurt.de
[*]The first two authors contributed equally to this work.
[4]http://www.6d-vision.com/lostandfounddataset

Along with these contributions, we introduce *Lost and Found*, the first dataset dedicated to visual lost cargo detection, to push forward research on these typically underrepresented but critical events. In our detailed evaluation, we introduce suitable metrics derived from related computer vision problems with the application focus in mind.

The paper is organized as follows: Section II lists relevant related work concerning lost cargo detection. The proposed and baseline methods are explained in Section III. Our new dataset, an extensive evaluation of the presented methods and a discussion of the results are covered in Section IV. The final section comprises conclusions and future work.

## II. RELATED WORK

There exists a significant amount of literature on obstacle detection in general, spanning a variety of application areas. However, the literature focusing on detection of small obstacles on the road is quite limited. Most relevant are camera-based methods for the detection and localization of generic obstacles in 3D space, using small-baseline stereo setups ($< 25$ cm) on autonomous cars.

Many obstacle detection schemes are based on the flat-world-assumption, modeling free-space or ground as a single planar surface and characterizing obstacles by their height-over-ground [6], [7], [8]. Geometric deviations from the reference plane can be estimated either from a precomputed point cloud, directly from image data [9], or via mode extraction from the v-disparity histogram on multiple scales [10]. To become independent of the often violated flat-world-assumption, more sophisticated ground profile models have been introduced, from piece-wise planar longitudinal profiles [11] to clothoids [12] and splines [13]. Also, parameter-free ground profile models have been investigated using mutliple filter steps and adaptive thresholding in the v-disparity domain [14].

The survey in [15] presents an overview of several stereo-based generic obstacle detection approaches that have proven to perform very well in practice. The methods are grouped into different obstacle representation categories and include Stixels [5], Digital Elevation Maps (DEM) [16] and geometric point clusters [17], [18]. Notably, all of the methods rely on precomputed stereo disparity maps. We select the Stixel method [5] as well as the point clustering method of [17], [18] to serve as baselines during our experimental evaluation.

The Stixel algorithm distinguishes between a global ground surface model and a set of rectangular vertical obstacle segments, providing a compact and robust representation of the 3D scene. In [17], [18], the geometric relation between 3D points is used to detect and cluster obstacle points.

The above methods yield robust detection results based on generic geometric criteria and perform best for detecting medium-sized objects at close to medium range. Detection performance and localization accuracy drop with increasing distance and decreasing object sizes.

Our present work builds on the approach presented in [4], originally devised for high sensitivity in long range detection tasks. Obstacle and free-space hypotheses are tested against each other using local plane models, optimized directly on the underlying image data.

Several works use appearance cues in addition to geometric information to improve generic obstacle detection. Detection of lost cargo can also be mapped to the task of anomaly detection [19]. There, every patch that deviates from the learned road appearance is considered a potential hazard. Due to the lack of 3D information, this method also triggers on patches of harmless flat objects. In [20], color and texture cues are used for stereo-based obstacle detection incorporating the cues into the Stixel framework of [5]. For off-road driving, basic deep learning techniques have been employed to detect driveable regions [21]. There, stereo-based obstacle detection acts as a supervisor for collecting training samples in the short range while the classifier yields predictions for the long range. A similar idea has been pursued in [22], however, instead of learning-based predictions, a spectral clustering of superpixels is applied for the long-range prediction.

Overall, the - by definition - unknown type of the objects to be detected makes the direct application of common machine learning-based approaches difficult.

## III. METHODS

### A. Direct Planar Hypothesis Testing (PHT)

In this work we build upon and extend the geometric obstacle detection approach proposed in [4]. We refer to this method as *Direct Planar Hypothesis Testing (PHT)*, since the detection task is formulated as a statistical hypothesis testing problem on the image data.

Free-space is represented by the null hypothesis $\mathcal{H}_f$, while obstacles correspond to the alternative hypothesis $\mathcal{H}_o$. The hypotheses are characterized by constraints on the orientations of local 3D plane models, each plane being defined by a parameter vector $\vec{\theta}$ comprised of the normal vector $\vec{n}$ and the normal distance $D$ from the origin: $\vec{\theta} = (n_X, n_Y, n_Z, D)^T$. The parameter spaces of $\mathcal{H}_f$ and $\mathcal{H}_o$ are bounded by the maximum allowed deviations $\widetilde{\varphi}_i = \widetilde{\varphi}_{\{f,o\}}$ of the normal vectors from their reference orientation (see Fig. 2a).

For each local image patch an independent Generalized Likelihood Ratio Test (GLRT) is formulated using the Maximum Likelihood Estimates (MLE) $\hat{\vec{\theta}}_i$ of the respective hypothesis parameter vector. The decision for $\mathcal{H}_o$ is then taken based on the threshold $\gamma$ according to the criterion

$$\ln\left(p(\vec{\mathcal{I}}; \hat{\vec{\theta}}_o, \mathcal{H}_o)\right) - \ln\left(p(\vec{\mathcal{I}}; \hat{\vec{\theta}}_f, \mathcal{H}_f)\right) > \ln(\gamma). \quad (1)$$

The likelihood terms $p(.)$ are derived directly from a statistical model of the stereo image data $\vec{\mathcal{I}}$, where the left and right intensity values $I_l(\vec{x})$ and $I_r(\vec{x})$ represent noisy samples of the observed image signal $f$ at position $\vec{x} = (x, y)^T$, with

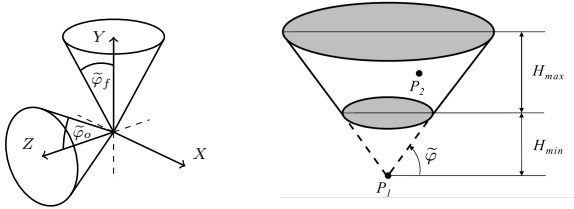

(a) Direct Planar Hypothesis Testing (PHT) [4]: The angles $\widetilde{\varphi}_f$ and $\widetilde{\varphi}_o$ constrain the allowed plane normal orientations. The Z axis represents the optical axis of the camera

(b) Point Compatibility (PC) [17], [18]: Any point $P_2$ lying within the given truncated cone based at point $P_1$ is labeled as obstacle and the points are said to be compatible, i.e. are part of an obstacle cluster

Fig. 2. Geometric models for point-wise obstacle detection criteria

local bias $\alpha(\vec{x})$ and zero-mean noise samples $\eta(\vec{x})$:

$$I_l(\vec{x}) = f(\vec{x}) + \alpha_l(\vec{x}) + \eta(\vec{x}) \quad (2)$$
$$I_r\left(W(\vec{x}, \vec{\theta})\right) = f(\vec{x}) + \alpha_r(\vec{x}) + \eta(\vec{x}). \quad (3)$$

The warp $W$ transforms the image coordinates $\vec{x}$ from the left to the right image, according to the plane model of the true hypothesis and the camera parameters $\mathbf{P} = \mathbf{K}[\mathbf{R}|\vec{t}]$. For the models used in [4] this warp represents a multiplication by the plane-induced homography $\mathbf{H} = \mathbf{K}\left(\mathbf{R} - \frac{1}{D}\vec{t}\vec{n}^T\right)\mathbf{K}^{-1}$.

Each hypothesis' contribution to (1) then reduces to a sum $F$ of pixel-wise residuals over the local patch area $\Omega$

$$F_i = \sum_{\vec{x} \in \Omega} \rho\left(I_r\left(W(\vec{x}, \vec{\theta}_i)\right) - f(\vec{x})\right). \quad (4)$$

The loss function $\rho$ depends on the assumed noise model, and the unknown image signal $f$ is being estimated as the mean of the concurrently realigned input images.
Finding the MLE $\hat{\vec{\theta}}_i$ corresponds to the non-linear optimization problem with simple bound constraints

$$\hat{\vec{\theta}}_i \leftarrow \arg\min_{\vec{\theta}_i}(F_i) \quad s.t. \quad |\varphi_i| \leq \widetilde{\varphi}_i. \quad (5)$$

The optimization problem is solved iteratively using a projected Levenberg-Marquardt method, similar to the local approach proposed in [23].

Note that only reliable decisions are reported, i.e. for patches in sufficiently textured image areas, where the minimum eigenvalue of the approximate Hessian is sufficiently large during optimization.

### B. Fast Direct Planar Hypothesis Testing (FPHT)

The PHT method provides high flexibility in terms of both model parameters and multi-camera configurations. However, for calibrated stereo cameras a simplified parametrization can be utilized, reducing the number of free parameters and the complexity of the optimization problem.

Thus we propose a Fast Direct Planar Hypothesis Testing (FPHT) method that exploits such a reparametrization, resulting in a computational speed-up of one order of magnitude without sacrificing detection performance in practice. The proposed reparametrization is based on considering

- rectified stereo image pairs and
- plane models without yaw or roll angles, i.e. $n_X = 0$.

Under these assumptions, computation of the warp $W$ is simplified significantly, since a plane with $n_X = 0$ can be represented by a line in stereo disparity space [11]:

$$W(\vec{x}, \vec{\theta}) = \begin{pmatrix} x - d \\ y \end{pmatrix} = \begin{pmatrix} x - (a\bar{y} + b) \\ y \end{pmatrix}. \quad (6)$$

Disparity is denoted by $d$, while $\bar{y} = \frac{y_c - y}{h/2}$ represents the normalized vertical image coordinate, with the patch center position $y_c$ and height $h$.

The new parameter vector $\vec{\theta}^* = (a, b)^T$ consists only of the disparity slope $a$ and offset $b$, which directly relate to the 3D plane parameters as

$$a = -n_Y \frac{f_x}{f_y} \frac{B}{D}, \quad b = -\frac{B}{D}\left(n_Y(y_c - y_0)\frac{f_x}{f_y} + n_Z f_x\right) \quad (7)$$

where the camera's focal lengths, vertical principal point and baseline are denoted by $f_x, f_y, y_0$ and $B$, respectively.

In contrast to the PHT method, where the parameters are bounded by globally valid box constraints, we have to consider the fact that two planes with *the same* slope angle $\varphi_i$ in 3D space, but with different normal distances $D$ to the camera, will in general have different disparity offsets $b$ *as well as* different slopes $a$ in disparity space. It is therefore not possible to specify independent global bounds on $a$ and $b$. Instead, we formulate the constraints by plugging the original bounds on the normal vector $\vec{n}$ into the linear relation

$$a = \frac{b}{(y_0 - y_c) + f_y \frac{n_Z}{n_Y}} = b \cdot const. \quad (8)$$

If an optimization step violates the bound and results in an invalid configuration of $a$ and $b$, both values are warped back onto the bounding line via vector projection.

As our principal motivation for FPHT is efficiency, we skip the estimation of the true image signal as in the PHT method and instead use the left image sample as the reference estimate. For both PHT and FPHT we use a Gaussian noise model, i.e. a squared error for the loss $\rho$.

### C. Point Compatibility (PC)

As a first baseline we use the Point Compatibility (PC) approach proposed in [17] and successfully applied in [18]. This geometric obstacle detection method is based on the relative positions of pairs of points in 3D space. Placing a truncated cone on a point $P_1$ as shown in Fig. 2b, any point $P_2$ lying within that cone is labeled as obstacle and said to be *compatible* with $P_1$. The cone is defined by the maximum slope angle $\widetilde{\varphi}$, the minimum relevant obstacle height $H_{min}$ and the maximum connection height threshold $H_{max}$.

All points of a precomputed stereo disparity map are tested in this way by traversing the pixels from bottom left to top right. The truncated cones are projected back onto the image plane and the points within the resulting trapezium are labeled accordingly. The algorithm not only provides a pixel-wise obstacle labeling but at the same time performs a meaningful clustering of compatible obstacle points [18].

Similar to PHT and FPHT, the PC approach does not depend on any global surface or road model due to its relative geometric decision criterion. However, it does depend directly on the quality of the underlying point cloud.

### D. Stixels

Furthermore, we use the Stixel approach of [5] as a baseline. Stixels provide a compact and robust description of 3D scenes, especially in man-made environments with predominantly horizontal and vertical structures. The algorithm distinguishes between a global ground surface model and a set of vertical obstacle segments of variable height. The segmentation task is based directly on a precomputed stereo disparity map and is performed column-wise in an optimal way via dynamic programming. The algorithm makes use of an estimated road model, a B-spline model as in [13], and incorporates further features such as ordering and gravitational constraints.

The results of the Stixel computation depend both on the quality of the disparity map as well as the estimated road model.

### E. Mid-level Representation: Cluster-Stixels (CStix)

Inspired by the compactness and flexibility of the Stixel representation, we present a corresponding extension for point-wise obstacle detection approaches such as PHT, FPHT and PC. Our aim is to create an obstacle representation similar to the Stixel algorithm, reducing the amount of output data and at the same time increasing robustness. Additionally, interpolating the sparse point clouds can even increase detection performance.

The proposed approach does not perform column-wise optimization like the actual Stixel algorithm, but consists of a clustering and a splitting step (cf. Alg. 1).

*1) Clustering:* In the first step, density-based geometric point clustering is performed via a modified DBSCAN algorithm [24]. We approximate the circular point neighborhood regions of the original algorithm by rectangles for efficient data access using bulk-loaded R-trees [25], [26]. The orientation of the rectangular neighborhood regions is aligned to the viewing rays of the camera.

Furthermore, we introduce several distance-adaptive modifications, considering the characteristics of stereo-based point clouds. The sizes of neighborhood regions are adapted to the points' distances, according to the estimated disparity noise. Also, the minimum number of cluster points is scaled with the distance from the camera, where the scaling formula is similar to the coordinate scaling in [12]: $minPts = minPts_0 + k \cdot \frac{f_x}{Z}$.

The adaptive DBSCAN algorithm allows for the use of meaningful clustering parameters, combining real-world dimensions and disparity uncertainty, and avoids discretization artifacts typical of e.g. scaled grid maps.

Note that for the PC approach the clustering step is omitted, since the detection algorithm itself already provides a set of meaningful clusters.

*2) Splitting:* After the clustering phase, each cluster is split horizontally and vertically into a set of Stixel-like vertical boxes. The horizontal splitting step strictly enforces a fixed box width to ensure the characteristic Stixel layout. The optional vertical splitting step takes the precomputed disparity map into account and performs recursive splits only as long as the disparity variance of a Stixel box exceeds a certain threshold.

---
**Algorithm 1** Mid-level representation: Cluster-Stixels
---
**Input**
  - list of obstacle points $\vec{P}$, e.g. from PHT, FPHT, PC
  - dense or sparse disparity map $\mathcal{D}$

**Output**
  - list of obstacle Cluster-Stixels $\overrightarrow{CStix}$

**Algorithm**
1: **function** MIDLEVELREP( $\mathcal{D}$ )
2:     $\vec{C} \leftarrow$ ADAPTIVEDBSCAN( $\vec{P}$ )  ▷ Compute list of obstacle clusters $\vec{C}$ with associated points
3:     $\overrightarrow{CStix} \leftarrow$ SPLITANDFIT( $\vec{C}$, $\mathcal{D}$ )  ▷ Split clusters $\vec{C}$ and fit Bounding Boxes (BB)
4:     **return** $\{\overrightarrow{CStix}\}$
5: **end function**

1: **function** SPLITANDFIT( $\vec{C}$, $\mathcal{D}$ )
2:     **for all** $C \in \vec{C}$ **do**
3:         $\overrightarrow{CStix} \stackrel{+}{\leftarrow}$ SPLITHORIZONTALLY( $C$, $width$ )  ▷ split horizontally and fit BB with fixed Stixel width
4:         $\overrightarrow{CStix} \stackrel{+}{\leftarrow}$ SPLITVERTICALLY( $\overrightarrow{CStix}$, $C$, $\mathcal{D}$ )  ▷ split vertically until disparity variance in BB is below threshold
5:     **end for**
6:     **return** $\{\overrightarrow{CStix}\}$
7: **end function**
---

## IV. EVALUATION

### A. Lost and Found Dataset

In order to evaluate the performance of small road obstacle detection approaches, we introduce a novel dataset with recordings from 13 different challenging street scenarios, featuring 37 different obstacle types. The selected scenarios contain particular challenges including irregular road profiles, far object distances, different road surface appearance and strong illumination changes. The objects are selected to reproduce a representative set that may actually appear on the road in practice (see Fig. 3). These objects vary in size and material, which are factors that define how hazardous an object may be for a self-driving vehicle in case the obstacle is placed within the driving corridor. Note that we currently treat some very flat objects (i.e. lower than 5 cm) as non-hazardous and thus do not take them into account in the results of Sect. IV-C.

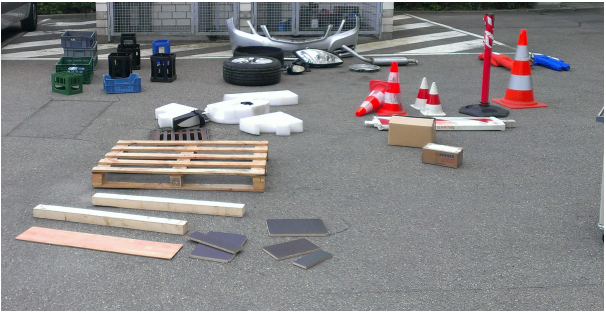

Fig. 3. Collection of objects included in the *Lost and Found* dataset

| Subset | Sequences | Frames | Locations | Objects |
|---|---|---|---|---|
| Train/Val | 51 | 1036 | 8 | 28 |
| Test | 61 | 1068 | 5 (5) | 35 (9) |

TABLE I

DETAILS ON THE DATASET SUBSETS. NUMBERS IN PARANTHESES REPRESENT UNSEEN TEST ITEMS NOT INCLUDED IN THE TRAINING SET

The *Lost and Found* dataset consist of a total of 112 video stereo sequences with coarse annotations of free-space areas and fine-grained annotations of the obstacles on the road. Annotations are provided for every 10th frame, giving a total of 2104 annotated frames (see Fig. 7). Each object is labeled with a unique ID, allowing for a later refinement into different subcategories (e.g. obstacle sizes).

The dataset is split into a Train/Validation subset and a Test subset. Each of these subsets consists of recordings in completely different surroundings, covering a similar number of video sequences, frames and objects (see Table I). The Test subset contains nine previously unseen objects that are not present at all in the Training/Validation subset. Further, the test scenarios can be considered to be more difficult than the training scenarios, owing e.g. to more complex road profile geometries.

The stereo camera setup features a baseline of 21 cm and a focal length of 2300 pixels, with spatial and radiometric resolutions of 2048×1024 pixels and 12 bits. While the dataset contains full color images, all methods considered in the present work make use only of the grayscale data.

To the best of our knowledge, this represents the first publicly available dataset with its main focus on the detection of small hazards and lost cargo on the road. We hope that this dataset supports further research on this critical topic for self-driving vehicles.

### B. Metrics

To quantitatively analyze the detection performance of the different approaches, pixel- and object-level metrics derived from related computer vision problems are defined while keeping the application focus in mind.

*1) Pixel-level Metric:* As a first metric we define a Receiver-Operator-Characteristic (ROC) curve that compares a pixel-wise True Positive Rate (TPR) over False Positive Rate (FPR). This ROC curve is generated by performing a parameter sweep and computing the convex hull over the results from all evaluated parameter configurations.

$$TPR = \frac{TP \cdot Sub^2 \cdot Dwn^2}{GT_{Obstacles}} \quad (9)$$

$$FPR = \frac{FP \cdot Sub^2 \cdot Dwn^2}{GT_{FreeSpace}} \quad (10)$$

$TP$ and $FP$ refer to the number of true and false pixel-wise predictions that a given method produces, which are evaluate with respect to the annotated image areas. $Sub$ and $Dwn$ are two scaling factors that compensate the subsampling and downsampling settings of some of the evaluated methods (cf. Sect. IV-C). Finally, $GT_{Obstacles}$ and $GT_{FreeSpace}$ are the total number of ground truth pixels labeled as obstacle or free-space respectively.

*2) Instance-level Metric:* The main drawback of the above described ROC curve is its bias toward object instances that cover large areas in the images. In order to overcome this disadvantage, we propose a second metric based on an instance-level variation of the Jaccard Index, known as instance Intersection over Union (iIoU) [27]. This second metric analyzes the instance-level Intersection (iInt) between mid-level predictions (Stixels) and pixel-wise ground truth annotations over the false positive Stixels per frame (FP/frame). A Stixel is defined as false positive if its overlap with the labeled free space area is larger than a given threshold (50% for the purpose of the evaluation presented in the next subsection).

$$iInt = \frac{iTP}{iTP + iFN} \quad (11)$$

$iTP$ and $iFN$ represent pixel-wise true positives and false negatives per instance.

### C. Results

*1) Experimental Setup:* All methods as described above are included in our experiments for evaluation. For all point-based methods PHT, FPHT and PC, by default we employ subsampling of stride two for higher computational efficiency. From our experience, this is a reasonable choice where no significant detection performance is being sacrificed. To further investigate the trade-off of efficiency and detection performance, we optionally scale down the images by an additional factor of two for even faster execution. In the following, results including this downscaling step will be denoted as *downsampled*.

We perform a parameter sweep over the principal parameters of each method, i.e.:

- PHT/FPHT: patch size $h \times w$, minimum eigenvalue for optimization, likelihood ratio decision threshold $\gamma$
- PC: maximum angle $\widetilde{\varphi}$, $H_{min}$, $H_{max}$
- Stixels: vertical cut costs used in dynamic programming

For PHT and FPHT, the exact bounds on the plane normals prove to be not too critical and we set them to $\widetilde{\varphi}_f = 25°$ and $\widetilde{\varphi}_o = 45°$. The parameter vectors are initialized using a dense disparity map, precomputed via Semi-Global Matching (SGM) [28]. The same disparity map provides the input to the PC and Stixel approaches.

For the additional computation of the mid-level Cluster-Stixels representation we use a fixed, manually optimized, set of parameters. In fact, these exact parameter values have a much lower impact on the final results than the parameters optimized in the sweep described above.

*2) Quantitative Results:* First, the primary methods (PHT, FPHT, PC and Stixels) are benchmarked using the Training/Validation subset and the described pixel-level ROC curve. For the purpose of this first evaluation, we perform a parameter sweep as described above. The best performing parameter configurations are then determined by computing the convex hulls over the TPR and FPR results (see Fig. 4). Note that the main purpose of this step is method-specific parameter optimization. Direct comparisons between the ROC curves of the various methods have to be approached with care, as e.g. the large-object-size bias mentioned earlier has to be considered.

Once the best performing parameter sets have been determined, a second pixel-level ROC curve is computed on the Test subset, including the primary approaches along with their corresponding Cluster-Stixels extensions (Fig. 5).

It can be observed that all methods except Stixels yield consistent results across Training and Test subsets. Stixels perform notably worse on the Test subset, as it contains rather challenging road profiles, where a failure of the road estimation module has fatal consequences for the FPR. Therefore, Stixel results are not shown within the axis scales of Fig. 5.

Note that the curves in Fig. 5 display a considerable gain provided by the proposed mid-level Cluster-Stixels representation.

Finally, we compare the Cluster-Stixels extensions of the proposed and baseline approaches using the defined instance-level metric. The results in Fig. 6 clearly show that the PHT/FPHT approaches significantly outperform both baselines, achieving $iInt$ values of approx. 0.4 at an average of 3 false positives per frame. Here the negative impact of downsampling becomes visible, since it mainly influences the smallest object instances at long distances, now being weighted equally by the metric.

Notably, it can be seen that the proposed FPHT method performs on a par with or even better than the original PHT variant. While the algorithmic core of PHT with downsampling takes approx. 500 ms to process a full image on a state-of-the-art GPU, our FPHT version requires only 50 ms.

*3) Qualitative Results:* Fig. 7 depicts qualitative results of the evaluated methods on three example scenarios from the Test subset. The left column shows a typical example of a small road hazard (bobby car) in a residential area. In this case, due to the flat road profile and the medium object size, all methods are able to successfully detect the object.

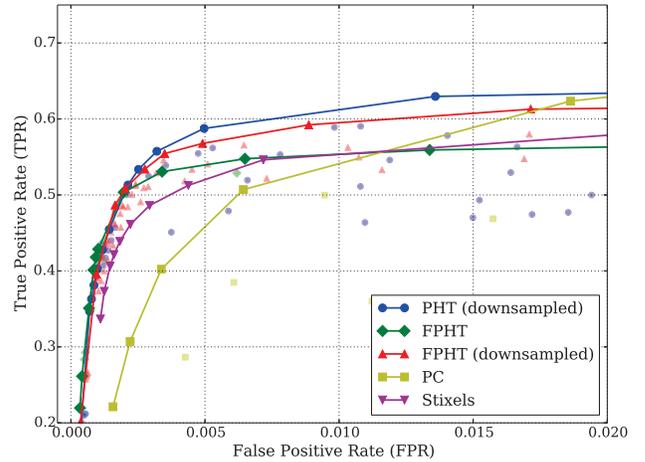

Fig. 4. Pixel-level: TPR over FPR (Training subset). Curves represent the convex hulls of the respective parameter sweep results

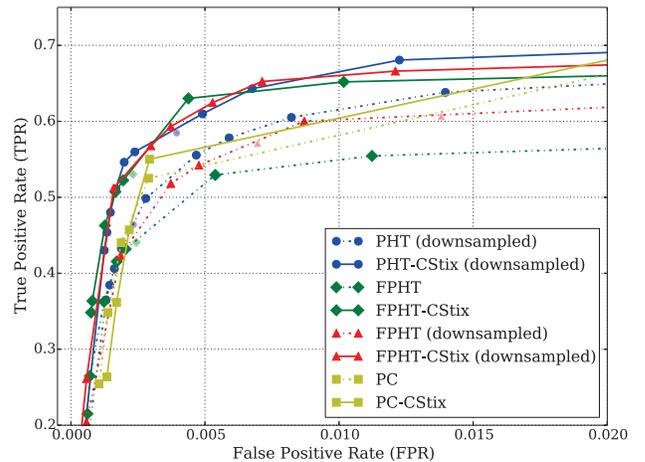

Fig. 5. Pixel-level: TPR over FPR (Test subset)

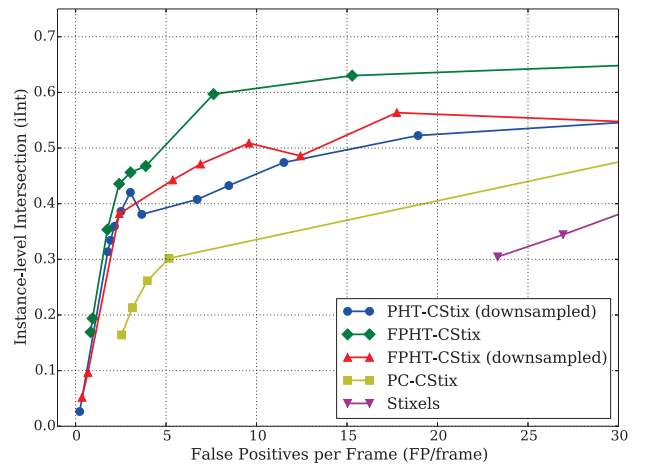

Fig. 6. Mid-level: iInt over FP/frame (Test subset)

In the middle column, an example with objects at large distances on a bumpy surface is shown. At such distances, the signal-to-noise ratio of the disparity measurements drops significantly, leading to a very low quality of the constructed 3D point cloud. Thus, neither the Stixel nor the PC approaches are able to detect the relevant objects in the scene. In contrast, the FPHT methods, which operate on the image data directly, still perform reasonably well at such large distances.

The scene in the rightmost column illustrates a rather challenging case for geometry-based obstacle detection approaches. A noticeable double kink in the longitudinal road profile would require an extremely accurate road model estimation for the Stixel method to be able to detect such small objects. While the PC and FPHT methods are invariant to such conditions, only FPHT succeeds in actually detecting the tire on the left side of the image. The tire simply appears to be not prominent enough for a PC-based detection. Considering the FPHT-CStix results, it can be seen that the detections cover larger portions of the obstacle than the FPHT results, which demonstrates the clustering step being a suitable compact representation. The second obstacle in the scene (square timber) is not detected by any of the methods due to its low profile.

Overall, from the observed qualitative results it can be concluded that the FPHT methods show the best performance for various obstacles and scenarios. The PC methods suffer from increased false positives rates, since noisy disparity measurements directly influence the results. This effect could possibly be reduced by applying sophisticated spatial and temporal disparity filtering methods. The qualitative results also confirm the Stixel method's dependency on a correctly estimated road profile.

## V. Conclusion

In this work, we have presented and evaluated an efficient stereo-based method for detecting small but critical road hazards, such as lost cargo, for self–driving vehicles. The approach extends previous work in this very relevant area, providing a significant gain in computational speed while at the same time outperforming all baseline methods. Additionally, the proposed mid-level representation Cluster-Stixels yields an extra gain in detection performance and robustness as well as a significant reduction in output complexity.

To allow for a detailed evaluation, we introduced the *Lost and Found* dataset, comprising over 2000 pixel-wise annotated stereo frames in a wide range of locations and road conditions and with generic objects of various types and sizes on the road. Based on the presented performance metrics, the experimental evaluation clearly demonstrates the efficacy of the proposed methods.

Future work may include a fusion of the considered geometric methods with an appearance-based approach using convolutional neural networks, where the *Lost and Found* dataset can be employed for training.

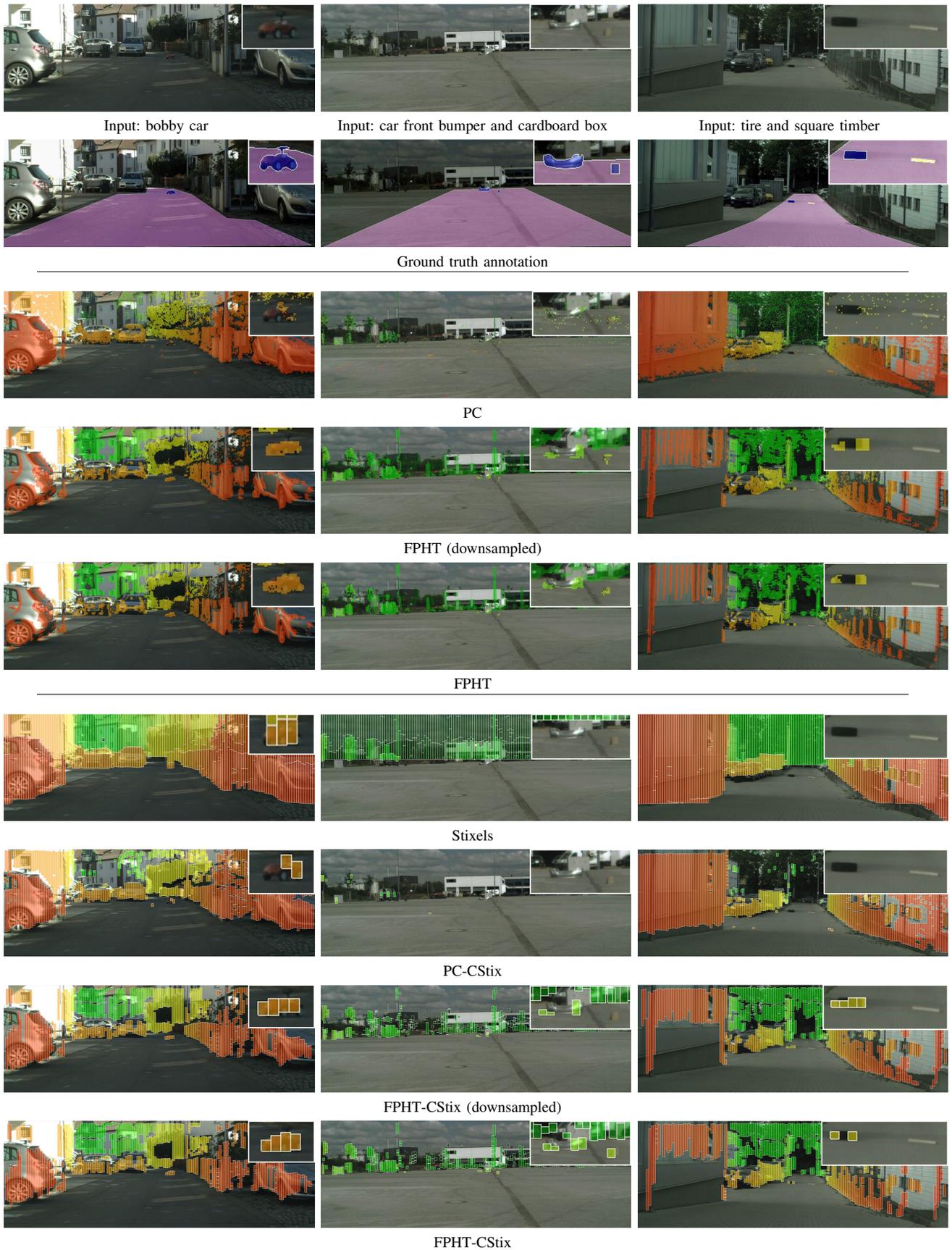

Fig. 7. Qualitative results of the evaluated methods. The top two rows show the left input image and the ground truth annotation, lower rows show pixle-wise and mid-level detections as overlay, color-coded by distance (red: near, green: far)